\documentclass[lettersize,journal]{IEEEtran}
\usepackage{amsmath,amsfonts}
\usepackage{algorithmic}
\usepackage{algorithm}
\usepackage{array}
\usepackage[caption=false,font=normalsize,labelfont=sf,textfont=sf]{subfig}
\usepackage{textcomp}
\usepackage{stfloats}
\usepackage{url}
\usepackage{verbatim}
\usepackage{graphicx}
\usepackage{cite}
\usepackage{optidef}
\usepackage{hyperref}
\usepackage{tabularx} 
\usepackage{booktabs}
\usepackage{caption}

\begin{document}


\title{Uncertainty in Supply Chain Digital Twins: A Quantum-Classical Hybrid Approach}


\author{
    \IEEEauthorblockN{
        A. Abdullah, 
F. R. Sandjaja,
A. Abdul Majeed,
G. Wickremasinghe,
K. Rafferty, 
        V. Sharma
    }\\
    \IEEEauthorblockA{
        \textit{School of Electronics, Electrical Engineering and Computer Science, Queen's University Belfast, UK}
    }
}

\maketitle

\section*{Abstract}
This study investigates uncertainty quantification (UQ) using quantum-classical hybrid machine learning (ML) models for applications in complex and dynamic fields, such as attaining resiliency in supply chain digital twins and financial risk assessment. Although quantum feature transformations have been integrated into ML models for complex data tasks, a gap exists in determining their impact on UQ within their hybrid architectures (quantum-classical approach). This work applies existing UQ techniques for different models within a hybrid framework, examining how quantum feature transformation affects uncertainty propagation. Increasing qubits from 4 to 16 shows varied model responsiveness to outlier detection (OD) samples, which is a critical factor for resilient decision-making in dynamic environments. This work shows how quantum computing techniques can transform data features for UQ, particularly when combined with classical methods. 



\section{\textbf{Introduction}}
\label{sec:introduction}
Unlike Industry 4.0, which focuses on automation and digitisation, Industry 5.0 focuses on human-centricity as one of its core values while digital twins (DT) and data analysis as core enablers \cite{xu2021industry}. Keeping humans in the loop for data analysis for dynamic environments regarding the supply chain is crucial in light of Industry 5.0\cite{modgil2023developing}\cite{breque2021industry}. Alongside this, informed, predictive, and resilient supply chain modelling has already earned importance during recent events \cite{ivanov2020predicting}. Researchers are focusing on the ability to predict these disruptive events using various methods, such as a one-class support vector machine algorithm \cite{ashraf2024disruption}. 

The evolution of predictive models such as advanced machine learning (ML) has benefited supply chain DTs \cite{baryannis2019supply}. These models undoubtedly show promising results in handling complexities of data compared to traditional methods, such as time series \cite{ni2020systematic}. This can be further enhanced with the help of quantum computing, which is now widely discussed in supply chain DTs for efficiently solving computationally expensive tasks \cite{jiang2022quantum}.

Developing methods for close-to-accurate predictions is essential. However, it is not the complete picture as the global network data can have inherent uncertainties, or the predictive models can introduce additional uncertainties to the predictions \cite{tyralis2024review}. And, not incorporating uncertainty analysis can sometimes cause optimistic predictions at the cost of substantial financial losses and operational problems \cite{blasco2024survey}. 

\begin{figure}[tbh]
\vspace{0.1cm}
    \centering
    \includegraphics[width=\columnwidth]{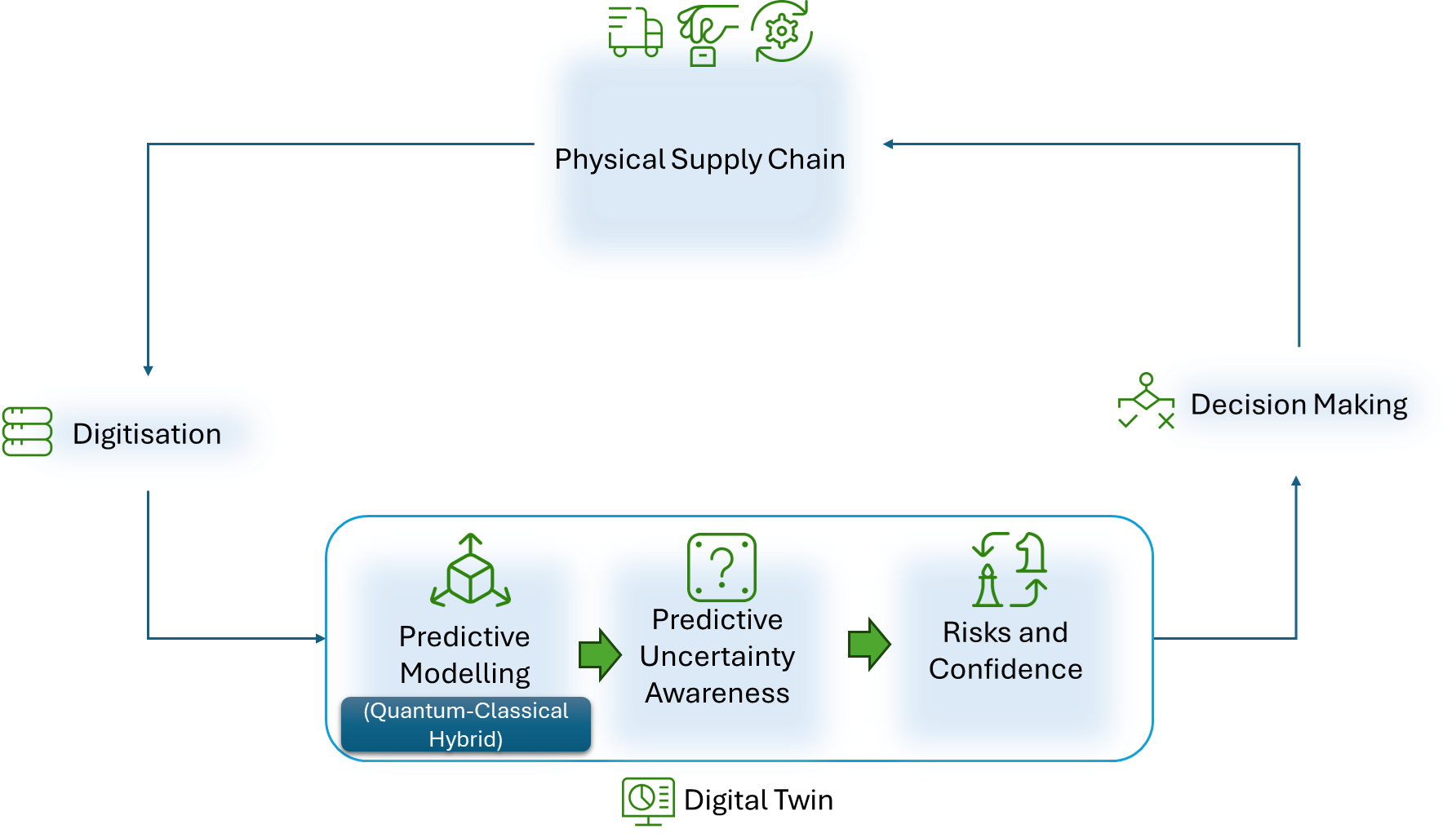}
    \caption{An exemplary overview of the supply chain with uncertainty-aware components for predictive UQ.}
    \label{fig:Outline}
    \footnotesize
    \vspace{\baselineskip}
\end{figure}

In this context, the ability to understand, analyse, and interpret uncertainties in DT predictions is vital. This could be crucial in effective decision-making and risk reduction in the supply chain domain. An exemplary overview of this process is given in Figure \ref{fig:Outline}.

In light of these developments and challenges, this work focuses on (1) quantifying predictive uncertainty in DTs based on quantum-classical hybrid predictive models, (2) assessing the propagation of uncertainty concerning various configurations of quantum circuits, and (3) analysing financial risks in a supply chain associated with different levels of uncertainty-aware models for providing insights into the economic value of supply chain. 
 
The area of uncertainty quantification (UQ) in predictive modelling has been widely explored (see Table~\ref{tab:relatedworktable}). However, UQ, in light of quantum-classical hybrid models, needs further investigation. This work seeks to analyse the effect of different levels of predictive uncertainties with the quantum-classical models. As quantum circuits are quite good at finding complexities within data, this work aims to help understand if these added dimensions affect the model's working with regard to uncertainty. We also focused on understanding the direct financial impact of uncertainty for these models in the supply chain DT.

The rest of this paper is structured as follows: Section 2 details the methodology, and Section 3 presents the performance of different models under various uncertainty conditions. Finally, Section 4 concludes with suggestions for future research directions.

\section{\textbf{Related work}}
\label{sec:related_work}
The predictive UQ and its impact have been studied for the classical ML and DL (Deep Learning) models. Over the years, quantum computing has been explored separately, and a hybrid approach has been used to leverage quantum properties such as superposition and entanglement. Quantum computers allow processing complex probability distributions more efficiently than classical systems\cite{schuld2021machine}, offering potential advantages in scenarios requiring rapid processing of uncertain data. For instance, Schuld et al. \cite{schuld2021machine} have demonstrated how quantum machine learning models could outperform classical models in learning and generalisation tasks under certain conditions, a foundational concept for applying quantum models to UQ. Cerezo et al. \cite{cerezo2021variational} have reviewed variational quantum algorithms (VQAs), which enable optimised parameter tuning within quantum circuits, enhancing the capacity of hybrid models to represent uncertainty effectively in high-dimensional contexts. 

A summarised literature for the most relevant works regarding the impact of the predictive UQ in various domains is given in Table \ref{tab:relatedworktable}. 

\begin{table*}[t]  
\centering
\small  
\caption{Relevant literature on predictive UQ and their impacts for classic and quantum ML models}
\label{tab:relatedworktable}
\begin{tabularx}{\textwidth}{X X X X}
\toprule
\textbf{Article} & \textbf{Approach} & \textbf{Parameters} & \textbf{Domain} \\
\midrule
\multicolumn{4}{l}{\textit{Classical machine learning/deep learning}} \\
\midrule
Choubineh et al. \cite{choubineh2023applying} & Monte Carlo (MC) dropout & Aleatoric and epistemic uncertainties & Subterranean fluid flow \\
Battula et al. \cite{battula2024uncertainty} & Bayesian inference, interval analysis & Sensor noise, model uncertainty & Digital twins for robotics, manufacturing \\
Siqueira et al. \cite{siqueira2024modelling} & Random forest with partial dependence & Environmental variability & Soil chemistry in Antarctica \\
Dutschmann et al. \cite{dutschmann2021evaluating} & High-variance leaves in random forests & Prediction uncertainty & Chemoinformatics \\
Nemani et al. \cite{nemani2023uncertainty} & Bayesian networks, neural ensemble methods & Gaussian process regression, Bayesian neural networks & Engineering design, health prognostics \\
\midrule
\multicolumn{4}{l}{\textit{Quantum machine learning}} \\
\midrule
Akhare et al. \cite{akhare2023diffhybrid} & DiffHybrid-UQ, Bayesian ensembles & Aleatoric and epistemic uncertainties & Physics-integrated SciML \\
Kim et al. \cite{kim2023quantum} & Quantum approximate Bayesian optimisation & Exploration and exploitation balancing & Mixed-integer optimisation \\
Nguyen et al. \cite{nguyen2022bayesian} & Bayesian quantum neural networks & Epistemic uncertainty, generalisation & Quantum machine learning \\
Park et al. \cite{park2023quantum} & Quantum conformal prediction & Calibration and error bars & General quantum ML models \\
Jahin et al. \cite{jahin2024triqxnet}  & TriQXNet with conformal prediction & Real-time data variability & Space weather forecasting \\
\bottomrule
\end{tabularx}
\end{table*}

Recently, Choubineh et al. \cite{choubineh2023applying} have employed Monte Carlo dropout in CNNs for subterranean fluid flow, quantifying aleatoric and epistemic uncertainties and improving reliability in high-dimensional regression predictions. Battula et al. \cite{battula2024uncertainty} have applied Bayesian inference and interval analysis to digital twins, focusing on real-time uncertainty control under sensor noise and model abstraction, targeting robotics and manufacturing. Siqueira et al. \cite{siqueira2024modelling} have used random forests with partial dependence for Antarctic soil chemistry prediction, controlling environmental variability factors like temperature and precipitation. Dutschmann et al. \cite{dutschmann2021evaluating} have used high-variance leaves in random forests to measure prediction uncertainty in chemoinformatics, specifically for molecular property predictions in drug design. Nemani et al. \cite{nemani2023uncertainty} have provided a tutorial on Gaussian process regression and Bayesian neural networks for uncertainty quantification in engineering and health prognostics, examining critical predictive factors in battery life and turbofan health.
Akhare et al. \cite{akhare2023diffhybrid} have introduced DiffHybrid-UQ, employing Bayesian ensembles to manage aleatoric and epistemic uncertainties in physics-integrated SciML, using Bayesian model averaging for efficient uncertainty propagation in PDE-based systems. Kim et al. \cite{kim2023quantum} have developed a dual-mixer quantum approximate Bayesian optimisation (QABOA) algorithm, enhancing exploration-exploitation balance in mixed-integer optimisation, using quantum walks and Grover mixers. Nguyen et al. \cite{nguyen2022bayesian} have leveraged Bayesian quantum neural networks (QNNs) to improve generalisation and epistemic uncertainty in quantum machine learning, demonstrating enhanced accuracy over frequentist QNNs. Park et al. \cite{park2023quantum} have proposed quantum conformal prediction (QCP) to calibrate predictive intervals in quantum ML, using a calibration dataset to ensure reliability under quantum noise. Jahin et al. \cite{jahin2024triqxnet} have presented TriQXNet, a hybrid classical-quantum framework for real-time disturbance storm-time (Dst) index forecasting in space weather, implementing conformal prediction for robust uncertainty quantification.

Recent research \cite{park2023quantum} has introduced quantum conformal prediction as a nonparametric framework for establishing reliable predictive intervals, particularly beneficial in noisy environments, though not directly implemented in the work performed in this paper but provides an inspiration towards this study. Lloyd et al. \cite{lloyd2020quantum} have demonstrated that larger quantum feature spaces, made possible by increasing qubit counts, can improve the capacity for complex data representation and generalisation, suggesting a pathway to enhanced robustness and more reliable outlier detection in data-intensive applications. 
Additionally, KACQ-DCNN \cite{jahin2024kacqdcnnuncertaintyawareinterpretablekolmogorovarnold} and TriQXNet \cite{jahin2024triqxnet} introduced uncertainty-aware models that enhance interpretability and reliability in healthcare and solar wind applications.  The work on DiffHybrid-UQ \cite{akhare2023diffhybrid} illustrates how differentiable neural models can improve UQ in hybrid architectures. 

The work presented in this paper examines how quantum-based data transformations influence uncertainty measurements in hybrid systems, particularly when varying the number of qubits. Different statistical methods have been applied to compare classical and quantum-enhanced approaches, which show practical applications in supply chain DTs.


\section{Methodology}
\label{sec:methodology}
\begin{figure*}[tbh]
\vspace{0.1cm}
    \centering
    \includegraphics[width=\textwidth]{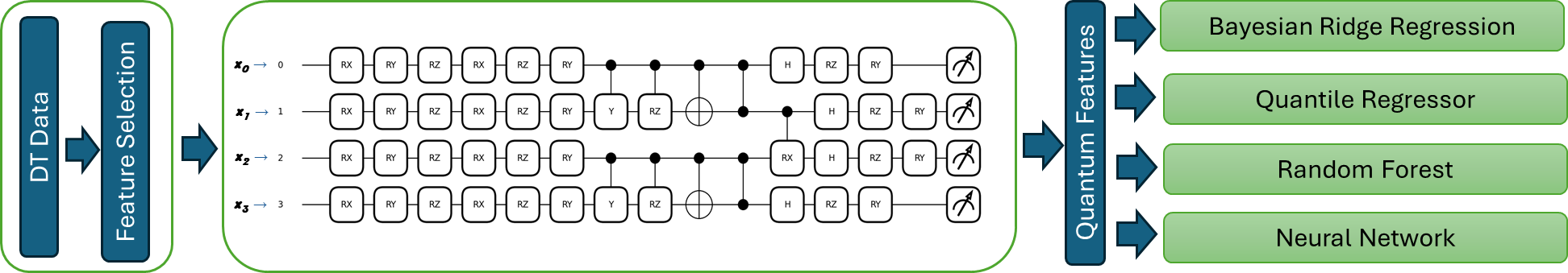}
    \caption{A detailed layout for the quantum-classical hybrid setup, including data pre-processing, feature selection, and quantum circuit configuration for UQ. }
    \label{fig:HybridQ_Outline}
    \footnotesize
    \vspace{\baselineskip}
\end{figure*}

After pre-processing, the synthesised DT data is reduced to the most important features and then fed into a customised quantum circuit. These quantum-processed outputs are then given to models such as Bayesian ridge regression (BRR), quantile regressor (QR), random forest (RF) and a neural network (NN). The choice of the models was based on the different nature of the predictive UQ methods of these models, which is the main goal of this study. The uncertainty for each model was calculated, and specific evaluations were performed. A detailed overview of the approach is given in Figure \ref{fig:HybridQ_Outline}. The base model for the analysis is the four qubits configuration that is compared to its classical version. However, further study is performed with a gradual increase of qubits in order to see the behaviour of predictive uncertainty. 

The details of the observations and complete methodology are as follows: 

\subsubsection{Data pre-processing and feature selection}
Synthetic data was generated for a supply chain DT that consists of features such as product demand, inventory levels and lead time. The distribution of the data, though synthetic, is provided in the supplementary files. Label encoding was performed for the categorical features, and then min-max scaling was performed. The target variable for this study is \textit{cost efficiency}, which is a score between 0 and 1. The recursive feature elimination (RFE) \cite{pedregosa2011scikit} method was used with a Gradient Boosting estimator for feature selection. The feature selection output was taken as 4, 6, 8, ..., and 16 features for different configurations of the quantum circuits. The configuration of 4 qubits can be seen in Figure \ref{fig:HybridQ_Outline}, while the rest of the configurations can be found in the supplementary files.

\subsubsection{Quantum feature engineering}
The input data \(\mathbf{x} = (x_1, x_2, \ldots, x_n) \in \mathbb{R}^n\) is encoded onto the quantum circuit using single-qubit rotations such as \( R_X(x_i) \), \( R_Y(x_i) \), and \( R_Z(x_i) \) gates \cite{nielsen2010quantum}. As shown in Figure \ref{fig:HybridQ_Outline}, input \(\mathbf{x} = (x_1, x_2, x_3, x_4)\)\ is being encoded onto to a 4 qubit circuit. These gates rotate the qubit states around the \(X\)-, \(Y\)-, and \(Z\)-axes, respectively, and are applied sequentially to each qubit state \( |q_i\rangle \) in the circuit. Specifically, each encoding gate is defined as follows: \( R_X(x_i) = e^{-i x_i X / 2} \), \( R_Y(x_i) = e^{-i x_i Y / 2} \), and \( R_Z(x_i) = e^{-i x_i Z / 2} \), where \( X \), \( Y \), and \( Z \) denote the Pauli matrices.

The circuit incorporates both single-qubit and multi-qubit operations. In each layer, parameterised single-qubit rotations \( R_X(\theta_i^{(X)}) \), \( R_Y(\theta_i^{(Y)}) \), and \( R_Z(\theta_i^{(Z)}) \) are applied to each qubit \( |q_i\rangle \), where the parameters \(\theta_i^{(X)}\), \(\theta_i^{(Y)}\), and \(\theta_i^{(Z)}\) are tunable values adjusted during the optimisation process.

For entangling operations, Controlled-Y (CY) gates, denoted by \( \text{CY}_{ij} = |0\rangle\langle0| \otimes I + |1\rangle\langle1| \otimes Y \), are used between pairs of qubits \( (|q_i\rangle, |q_j\rangle) \). The CY gate pairs are extended for larger circuit sizes (up to 16 qubits), following the specified configuration. Additionally, Controlled-RZ gates \( \text{CRZ}_{ij}(\phi) = |0\rangle\langle0| \otimes I + |1\rangle\langle1| \otimes R_Z(\phi) \) and CNOT gates \( \text{CNOT}_{ij} = |0\rangle\langle0| \otimes I + |1\rangle\langle1| \otimes X \) are used between adjacent qubits to further enhance entanglement. Explicit CZ gates \( \text{CZ}_{ij} = |0\rangle\langle0| \otimes I + |1\rangle\langle1| \otimes Z \) are also included to provide conditional phase flips between neighbouring qubits.

The circuit incorporates Hadamard gates, represented as \( H = \frac{1}{\sqrt{2}} \begin{pmatrix} 1 & 1 \\ 1 & -1 \end{pmatrix} \), which are applied to all qubits to create superposition states. These are followed by additional rotations \( R_Z \) and \( R_Y \) to increase expressivity. The circuit concludes with a series of measurements, extracting the expectation values of Pauli-\( X \), Pauli-\( Y \), and Pauli-\( Z \) observables for each qubit. For each qubit state \( |q_i\rangle \), these are given by \( \langle \psi | X_i | \psi \rangle \), \( \langle \psi | Y_i | \psi \rangle \), and \( \langle \psi | Z_i | \psi \rangle \), providing a set of features capturing the linear and non-linear relationships present in the input data.

The optimisation process employs a gradient-based approach to minimise a custom cost function combining mutual information loss \cite{pedregosa2011scikit} with a diversity penalty. Mini-batch gradient descent \cite{goodfellow2016deep} is used to iteratively update the quantum circuit parameters. The cost function focuses on maximising the mutual information between the quantum features and the target values while penalising high correlation among features to encourage diversity. 


The circuit and optimisation are implemented using the PennyLane library \cite{bergholm2018pennylane}, and feature selection and mutual information regression are handled using scikit-learn \cite{pedregosa2011scikit}.

\subsubsection{Models} 
BRR is implemented using the Scikitlearn library \cite{pedregosa2011scikit}. The model assumes a probabilistic framework where both the weights and output variance are treated as random variables with their distributions. QR is performed using Scikitlearn \cite{pedregosa2011scikit} to estimate the 25th, 50th, and 75th percentiles of the target distribution, which provides a non-parametric method to model different quantiles \cite{pedregosa2011scikit} \cite{koenker2005quantile}. RF is employed as an ensemble technique using an RF Regressor from Scikitlearn \cite{pedregosa2011scikit} with 100 decision trees. The model is trained using bootstrapped subsets of the data, where each tree is built on a random subset of features and samples \cite{breiman2001random}. NN is implemented using TensorFlow \cite{kohavi1995study} with an Adam optimiser \cite{kingma2014adam} for 200 epochs. The architecture comprises two hidden layers with 128 and 64 neurons, respectively, each followed by a dropout layer with a rate of 0.5 to regularise the model \cite{srivastava2014dropout}. For all models, 5-fold cross-validation is used to prevent overfitting \cite{kohavi1995study}. These models with the same specifications,  coupled with the quantum feature transformation, are used as hybrid quantum Bayesian ridge regression (HQBRR), hybrid quantum quantile regression (HQQR), hybrid quantum random forest (HQRF), hybrid quantum neural network (HQNN).

The method for quantification of predictive uncertainty was different for each model, which is given as:
\paragraph{Bayesian ridge regression (BRR)}
Uncertainty in BRR is quantified by separating it into aleatoric and epistemic components. Aleatoric uncertainty is calculated as the inverse of the regularisation parameter, which reflects the inherent noise in the data. Epistemic uncertainty is derived from the variance of the posterior distribution over the regression coefficients. This is calculated by taking the product of the test data and the covariance matrix of the coefficients. The total variance combines both uncertainties as predictive uncertainty and given as \cite{bishop2006pattern}\cite{korbak2020uncertainty}: 

\begin{equation}
\sigma^2_{\text{aleatoric}} = \frac{1}{\alpha},
\end{equation}

\begin{equation}
\sigma^2_{\text{epistemic}} = X_{\text{test}} \Sigma X_{\text{test}}^T,
\end{equation}

and

\begin{equation}
\sigma^2_{\text{total}} = \sigma^2_{\text{aleatoric}} + \sigma^2_{\text{epistemic}}.
\end{equation}

\paragraph{Quantile regressor (QR)}
In QR, uncertainty is quantified by constructing prediction intervals for different quantiles (25th and 75th). The uncertainty is represented by the width of the interval between the upper and lower quantiles, which directly captures the variability in the target prediction, and given as\cite{pedregosa2011scikit} \cite{koenker2005quantile} :

\begin{equation}
\text{Uncertainty} = Q_{0.75}(X) - Q_{0.25}(X),
\end{equation}

where $Q_{\tau}(X)$ is the $\tau$-th quantile prediction for input $X$.

\paragraph{Random forest (RF)}
For the RF model, UQ is achieved by analysing the variance across predictions made by each decision tree in the ensemble. Predictive uncertainty is captured by computing the variance among the predictions from all trees for each test sample, and given as \cite{tyralis2024review} \cite{pedregosa2011scikit}: 

\begin{equation}
\sigma^2 = \frac{1}{T} \sum_{t=1}^T (f_t(X) - \bar{f}(X))^2,
\end{equation}

where $f_t(X)$ is the prediction of the t-th tree and $\bar{f}(X)$ is the mean prediction of all trees.

\paragraph{Neural network (NN)}
The NN employs Monte Carlo Dropout for UQ. During inference, dropout layers remain active, and 1,000 forward passes are performed. Predictive uncertainty is captured by calculating the variance of the predictions across these passes, reflecting the model's uncertainty due to limited data, and given as \cite{gal2016dropout}:

\begin{equation}
\sigma^2 = \frac{1}{T} \sum_{t=1}^T (f(X, \omega_t) - \bar{f}(X))^2,
\end{equation}

where $f(X, \omega_t)$ is the prediction with dropout mask, $\omega_t$, and $\bar{f}(X)$ is the mean prediction over all passes.

\subsubsection{Standard performance}
Standard performance metrics such as mean squared error (MSE), mean absolute error (MAE), and R-squared (R2) are calculated for each model to assess their predictive accuracy. 
\subsubsection{Uncertainty-error correlation}
The uncertainty-error correlation is determined by calculating the Pearson correlation coefficient between the predictive uncertainties and the absolute prediction errors. 
\subsubsection{Prediction interval coverage probability (PICP)}
PICP is calculated by determining the proportion of actual values within the dynamically computed lower and upper confidence bounds of the predicted values (lower bound and upper bound). This is achieved by counting the number of samples where the actual values fall within these bounds and dividing by the total number of samples \cite{khosravi2011comprehensive}, providing a quantitative measure of the model's interval reliability.

\subsubsection{Outlier detection (OD)}
OD is performed using two methods: a bound-based method and an uncertainty-based method. The bound-based method identifies OD samples where the actual values fall outside the predicted confidence intervals, while the uncertainty-based method detects OD samples based on dynamically computed thresholds using the Interquartile Range (IQR) \cite{Enderlein1987HawkinsDM} of predictive uncertainties. 

\subsubsection{Statistical tests}
Wilcoxon Signed-Rank Test \cite{woolson2005wilcoxon} was used to compare predictive errors and uncertainties between quantum and classical models for the 4 qubits base model configuration. The Kruskal-Wallis H Test \cite{ostertagova2014methodology} was applied to assess differences in these metrics across multiple qubits configurations (4, 6, 8, ..., 16), allowing for non-parametric comparisons.


\section{Results and Discussion}
\label{sec:Results and Discussion}
\begin{figure}[tbh]
\vspace{0.1cm}
    \centering
    \includegraphics[width=\columnwidth]{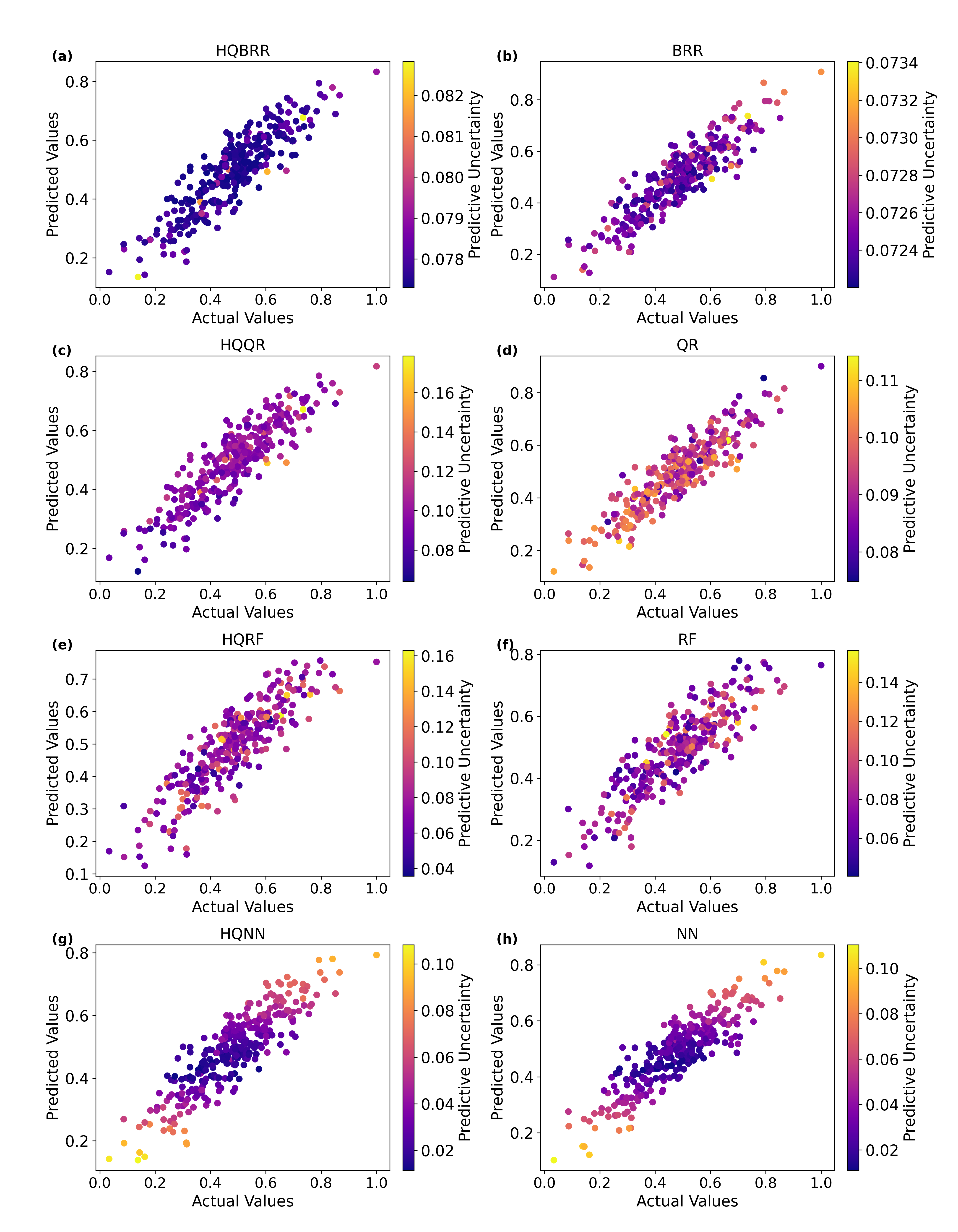}
  \caption{Actual vs. predicted comparison for classical and quantum-classical hybrid models. }
    \label{fig:comparison_actual_vs_predicted}
    \footnotesize
    \vspace{\baselineskip}
\end{figure}
This section provides the results and discussions based on the above-mentioned methodology using a synthetically generated dataset (provided as supplementary files).
\subsubsection{Comparative analysis of predictive uncertainty: classical vs. quantum hybrid models}
A comparison was conducted between classical and quantum-classical hybrid models trained on four selected features. Classical models (BRR, QR, RF, NN) directly used these features. In contrast, quantum-classical hybrid models applied them to 4 qubits quantum circuit, generating quantum-transformed features for the same predictive models. Performance metrics (R\textsuperscript{2} $\approx$ 0.79, RMSE $\approx$ 0.78, MAE $\approx$ 0.75) were comparable across both approaches, but for all of the models, differences were noted in uncertainty outcome as can be seen in Figure \ref{fig:comparison_actual_vs_predicted}.

\begin{figure}[tbh]
\vspace{0.1cm}
    \centering
    \includegraphics[width=\columnwidth]{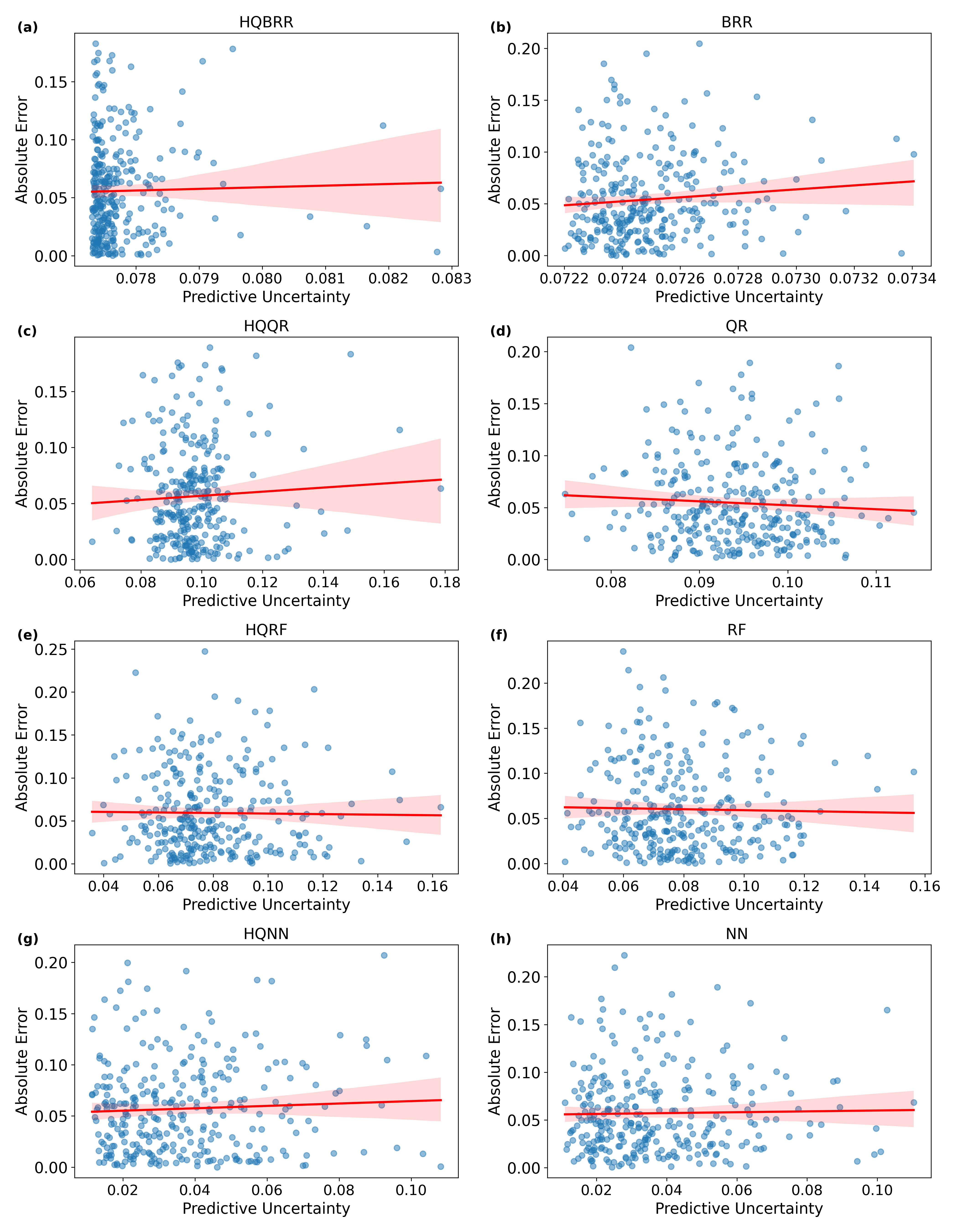}
    \caption{Uncertainty vs error for classic and hybrid models.}
    \label{fig:comparison_uncertainty_vs_error}
    \footnotesize
    \vspace{\baselineskip}
\end{figure}

\begin{table}[h]
    \centering
    \caption{Uncertainty and interval coverage metrics for classical and quantum-classical hybrid models (Corr - Correlation (Uncertainty vs Error), SII: Samples in interval, Int.: Interval, Unc.: Uncertainty), OD$_1$: Outlier detection interval and OD$_2$: Outlier detection uncertainty).}
    \label{tab:interval_uncertainty}
    \begin{tabular}{p{1.2cm}p{1.5cm}p{1.2cm}p{1cm}p{0.9cm}p{1cm}}
        \toprule
        \textbf{Model} & \textbf{Corr} & \textbf{SII} & \textbf{OD$_1$} & \textbf{PICP} & \textbf{OD$_2$} \\
        \midrule
        HQBRR & 0.0229 & 288 & 12 & 0.96 & 27 \\
        BRR & 0.0943 & 287 & 13 & 0.9567 & 11 \\
        HQQR & 0.0527 & 299 & 1 & 0.9967 & 15 \\
        QR & -0.0633 & 298 & 2 & 0.9933 & 1 \\
        HQRF & -0.0136 & 281 & 19 & 0.9367 & 7 \\
        RF & -0.0219 & 275 & 25 & 0.9167 & 3 \\
        HQNN & 0.0517 & 185 & 115 & 0.6167 & 10 \\
        NN & 0.0202 & 187 & 113 & 0.6233 & 9 \\
        \bottomrule
    \end{tabular}
\end{table}

This is further evident in the analysis of predictive uncertainty versus absolute error, which can be seen in Figure \ref{fig:comparison_uncertainty_vs_error}, where hybrid-quantum models, such as HQQR (Figure \ref{fig:comparison_uncertainty_vs_error} (c)), and  HQNN (Figure \ref{fig:comparison_uncertainty_vs_error} (g)) show a positive correlation, with slight increases in error as uncertainty rises when compared to their classical counterpart. However, for the classical model, BRR (Figure \ref{fig:comparison_uncertainty_vs_error} (b)) shows a stronger correlation as compared to its hybrid-quantum version (Figure \ref{fig:comparison_uncertainty_vs_error} (a)). The exact values of correlations can be seen in Table \ref{tab:interval_uncertainty}. These outcomes suggest that predictive uncertainty is differently informative for classical models apart from BRR. 

\begin{figure}[tbh]
\vspace{0.1cm}
    \centering
    \includegraphics[width=\columnwidth]{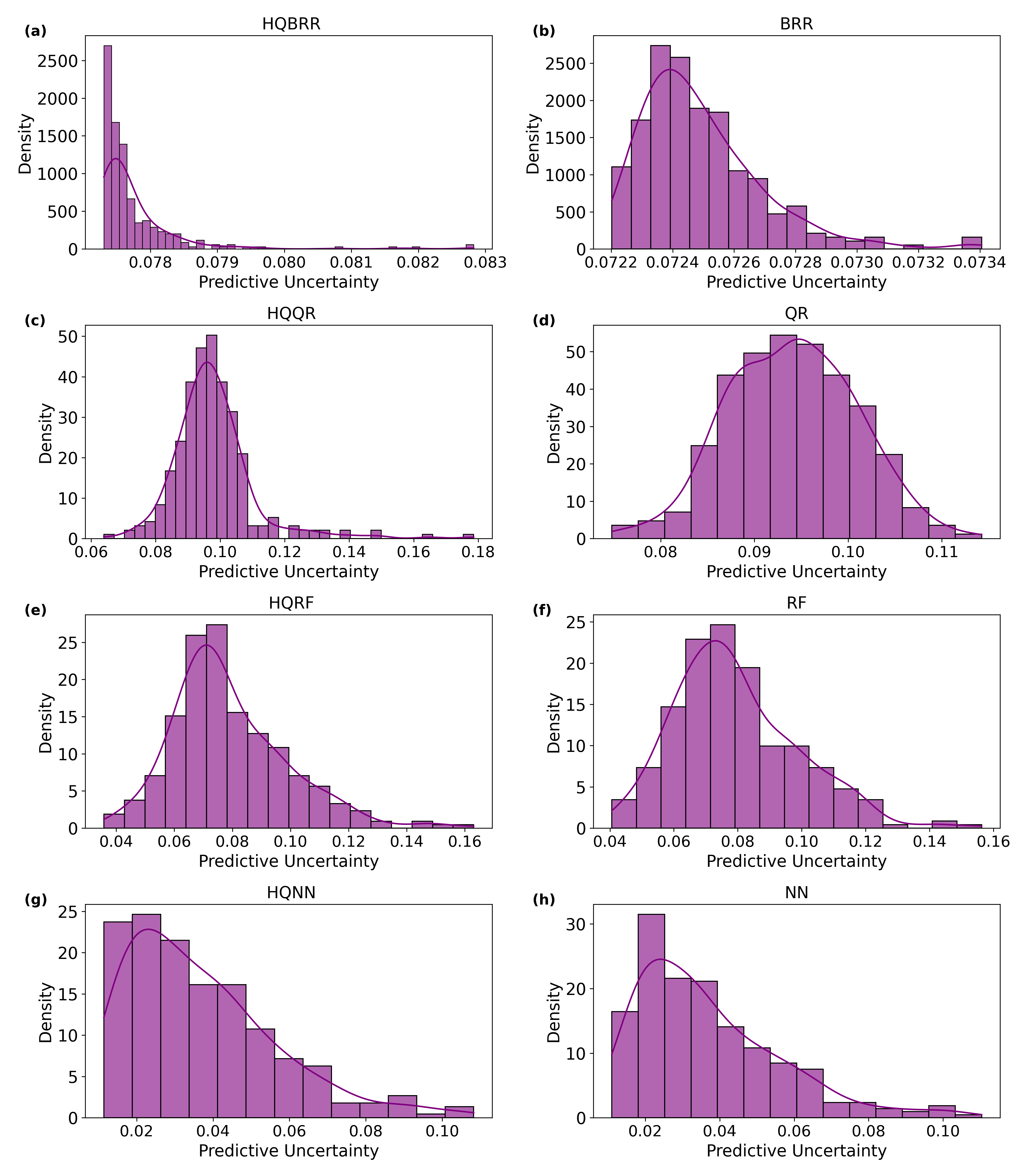}
    \caption{Distribution of predictive uncertainty for classical and quantum models.
}
    \label{fig:separate_uncertainty_distribution}
    \footnotesize
    \vspace{\baselineskip}
\end{figure}

The reason for these differences can be related to the nature of distribution of uncertainties as shown in Figure  \ref{fig:separate_uncertainty_distribution}, quantum models exhibited narrower uncertainty bounds compared to classical models (see Figure \ref{fig:comparison_uncertainty_vs_error}. Classical models like BRR (Figure \ref{fig:comparison_uncertainty_vs_error}(b)) and QR (Figure \ref{fig:comparison_uncertainty_vs_error}(d)) show broader uncertainty distributions, indicating greater variability, whereas quantum models demonstrated tighter distributions, suggesting reduced variance in the quantum feature space.

\begin{figure}[tbh]
\vspace{0.1cm}
    \centering
    \includegraphics[width=\columnwidth]{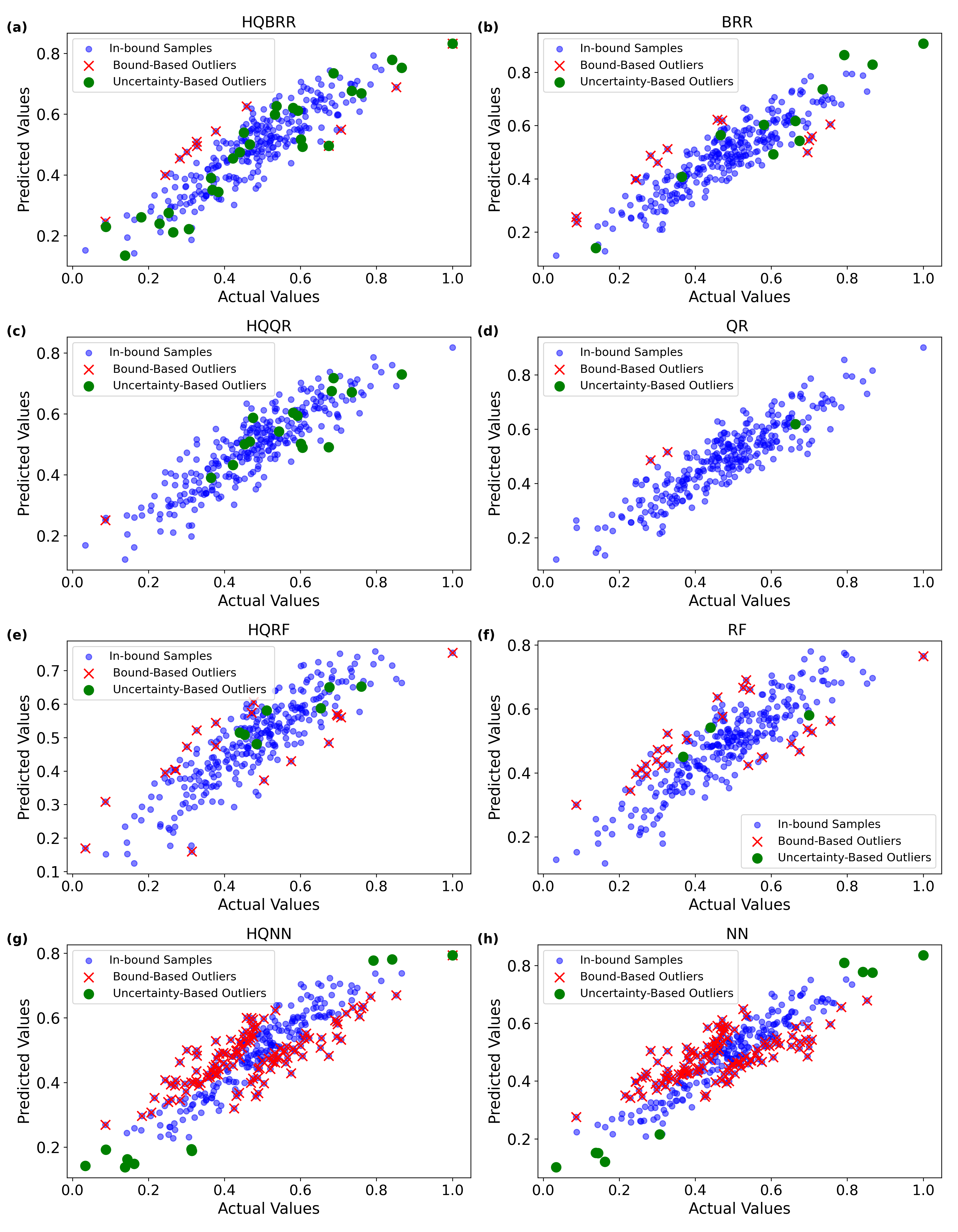}
    \caption{Outlier detection (OD) sample detection comparison between classical and quantum models using bound-based and uncertainty-based methods. }
    \label{fig:corrected_ood_samples.png}
    \footnotesize
    \vspace{\baselineskip}
\end{figure}

To further investigate the effects of these slight changes, the outliers related to uncertainties were calculated. OD was evaluated using bound-based and uncertainty-based methods. Quantum models identified more outlier samples due to distinct feature variance, particularly evident in Figure \ref{fig:corrected_ood_samples.png} and Table \ref{tab:interval_uncertainty}. It is observable that hybrid-quantum models show a higher number of outliers compared to classical models. Table  \ref{tab:interval_uncertainty} also highlights notable differences in interval reliability (PICP), especially for HQBRR and HQQR models.

Wilcoxon Signed-Rank Tests \cite{woolson2005wilcoxon} confirmed significant differences in predictive uncertainties between quantum and classical models for BRR (p = 6.08E-51) and QR (p = 9.33E-06) but not in predictive errors. The distinct uncertainty in HQBRR suggests enhanced sensitivity in UQ without affecting accuracy, indicating an enriched representation from quantum features in uncertainty modelling for BRR and QR.

\begin{figure}[tbh]
\vspace{0.1cm}
    \centering
    \includegraphics[width=\columnwidth]{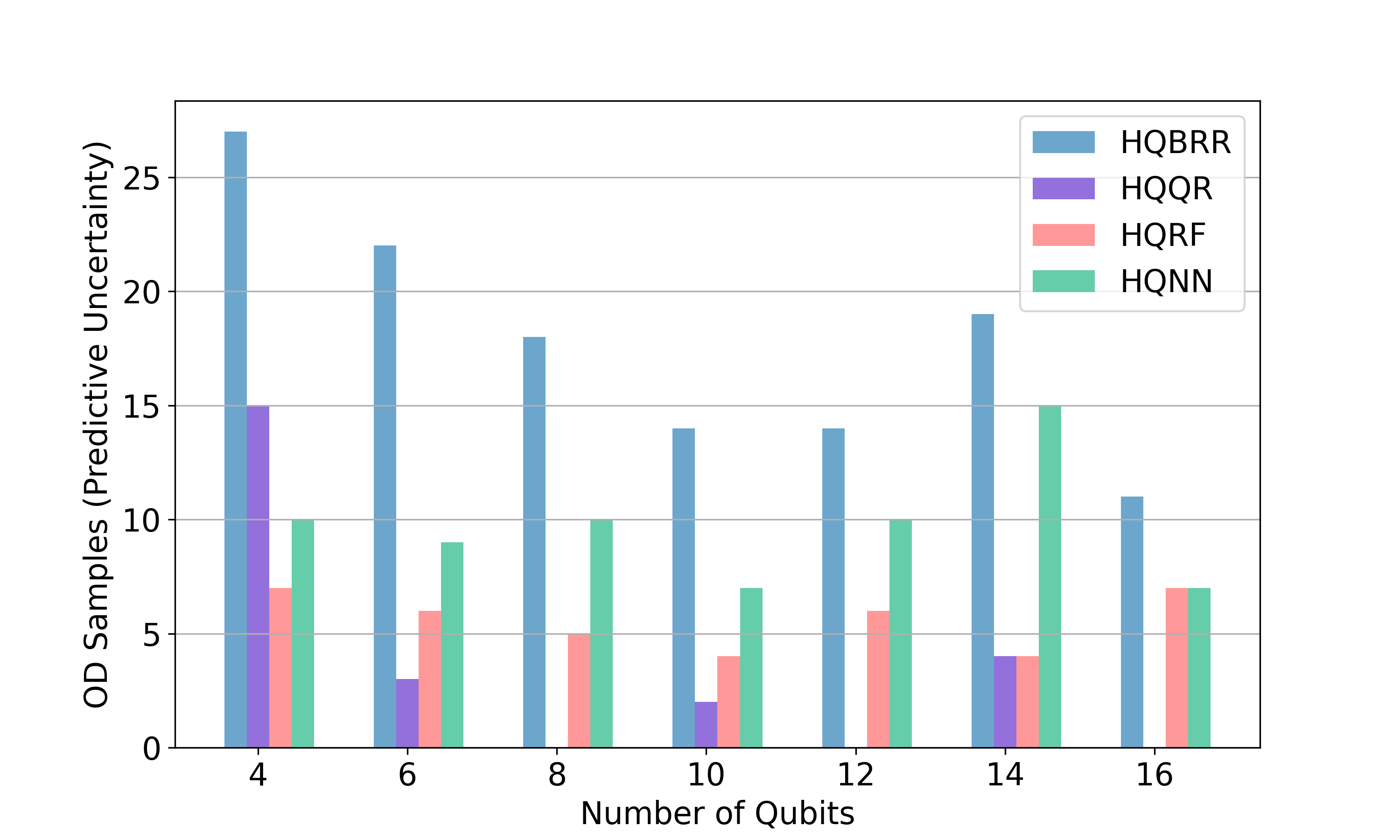}
    \caption{Impact of increasing qubit count on uncertainty propagation in quantum-classical hybrid models.}
    \label{fig:OOD_Samples_Uncertainty_Comparison}
    \footnotesize
    \vspace{\baselineskip}
\end{figure}

\subsubsection{Impact of increasing qubit count on uncertainty propagation}
Increasing qubit count (4--16) significantly impacts UQ in quantum-classical hybrid models. Kruskal-Wallis test \cite{woolson2005wilcoxon} confirms notable differences in uncertainty across configurations ($p < 0.05$) without affecting predictive errors ($p > 0.05$). Figure \ref{fig:OOD_Samples_Uncertainty_Comparison} shows a decline in OD samples with more qubits. HQBRR drops from 27 samples to 22, 18 for qubits 4th, 6th, 8th, 10th and 12th. At the 14th qubit, there is a rise and then a drop at 11 for qubit 16. HQQR shows a sharp reduction in outliers 15 to 3 after the 4th qubit. However, HQRF and HQNN do not display drastic changes. HQBRR and HQQR exhibit high sensitivity to qubit count in terms of outliers. However, this can not be generalised. These results indicate that higher qubit counts affect uncertainty modelling through feature representation and reduce OD detection, particularly in HQBRR and HQQR. The findings highlight the critical role of quantum dimensionality in refining UQ, with varied performance seen in higher qubit regimes.

\subsubsection{Translating uncertainty into financial impact} 
Given the significant differences in uncertainty across configurations but no direct correlation with prediction errors, the impact of uncertainty on the target variable `cost efficiency' was analysed. For this analysis, a Risk-Adjusted Cost Efficiency (RACE) is defined as a linear penalty formula and derived using \cite{keyCAPM}. $\beta$ values for risk-neutral as $\beta = 0$, low risk aversion as $\beta = 0.25$, moderate risk aversion as $\beta = 0.5$,  as high risk aversion $\beta = 0.75$ to very high risk aversions $\beta = 1.0$, were incorporated into the calculation of RACE, as

\begin{equation}
\text{RACE} = \text{Actual} - \beta \times \text{Uncertainty}.
\end{equation}

Additionally, the relative percentage difference (\(\Delta\)) between quantum and classical models was calculated as:

\begin{equation}
\Delta = \frac{\text{Quantum RACE Mean} - \text{Classical RACE Mean}}{\left|\text{Classical RACE Mean}\right|} \times 100.
\end{equation}

For uncertainties, direct standard deviations were used except for QR, where prediction intervals were standardised \cite{gareth2013introduction} to align with other models, ensuring comparability of scales.

\begin{figure}[tbh]
\vspace{0.1cm}
    \centering
    \includegraphics[width=\columnwidth]{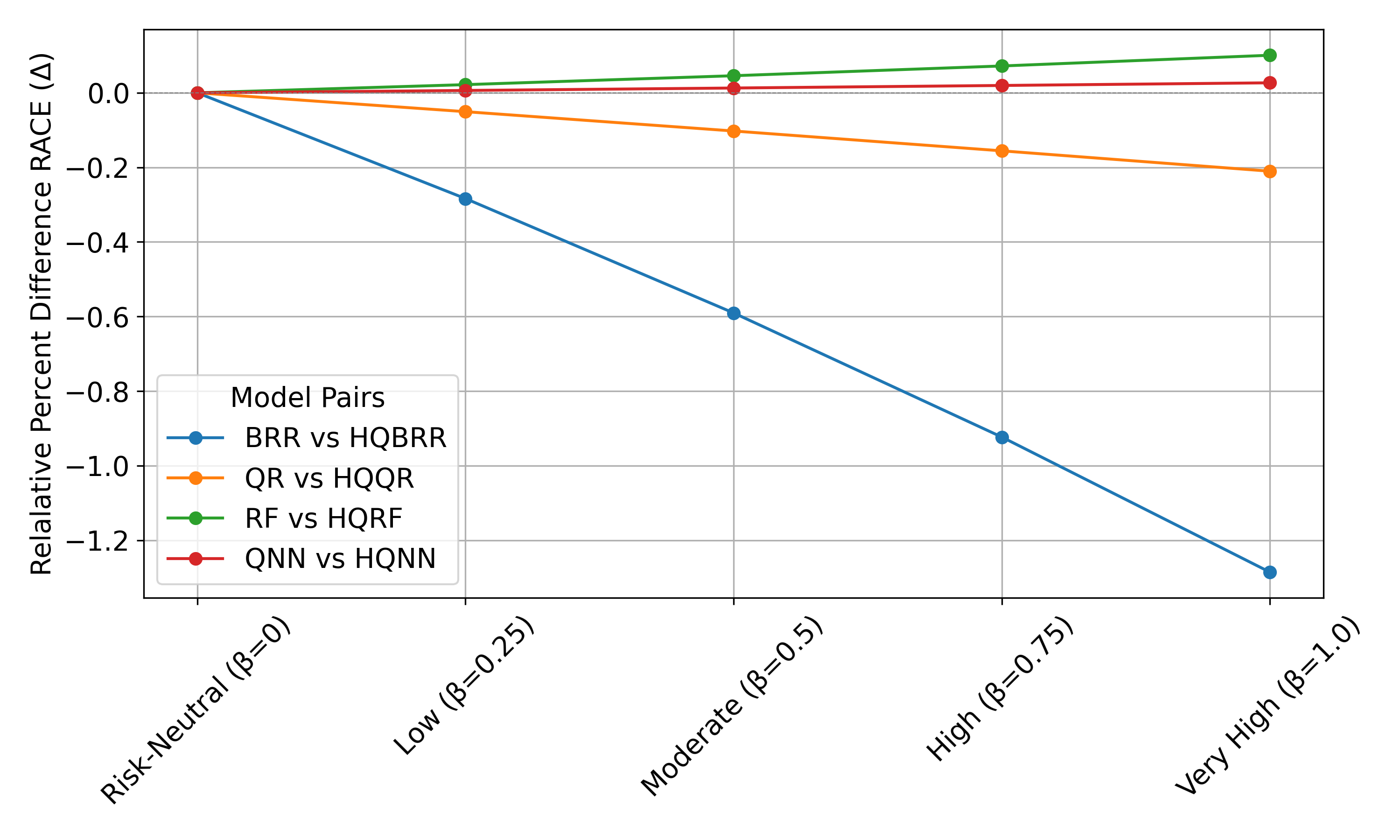}
  \caption{Comparison of risk-adjusted cost efficiency (RACE) for quantum and classical models across different risk aversion levels.}
    \label{fig:Quantum_vs_Classical_Models_Risk}
    \footnotesize
    \vspace{\baselineskip}
\end{figure}

\subsubsection{Quantum vs. classical comparison}

Figure \ref{fig:Quantum_vs_Classical_Models_Risk} shows differences in RACE across risk aversion levels for quantum and classical models. It is evident that the uncertainty impacts quantum-classical hybrid and classical models differently in terms of cost efficiency. BRR experiences a more significant decline in performance with quantum features under higher uncertainty, while models like RF show minimal difference. The RACE formula here applies a penalty for uncertainty. It can also be used to reward low uncertainty, showing an inverse effect. The cost efficiency score ranges from 0 to 1. However, in real-world financial scenarios, this could scale to millions of dollars, showing the impact of uncertainty on predictions.

\begin{figure}[tbh]
\vspace{0.1cm}
    \centering
    \includegraphics[width=\columnwidth]{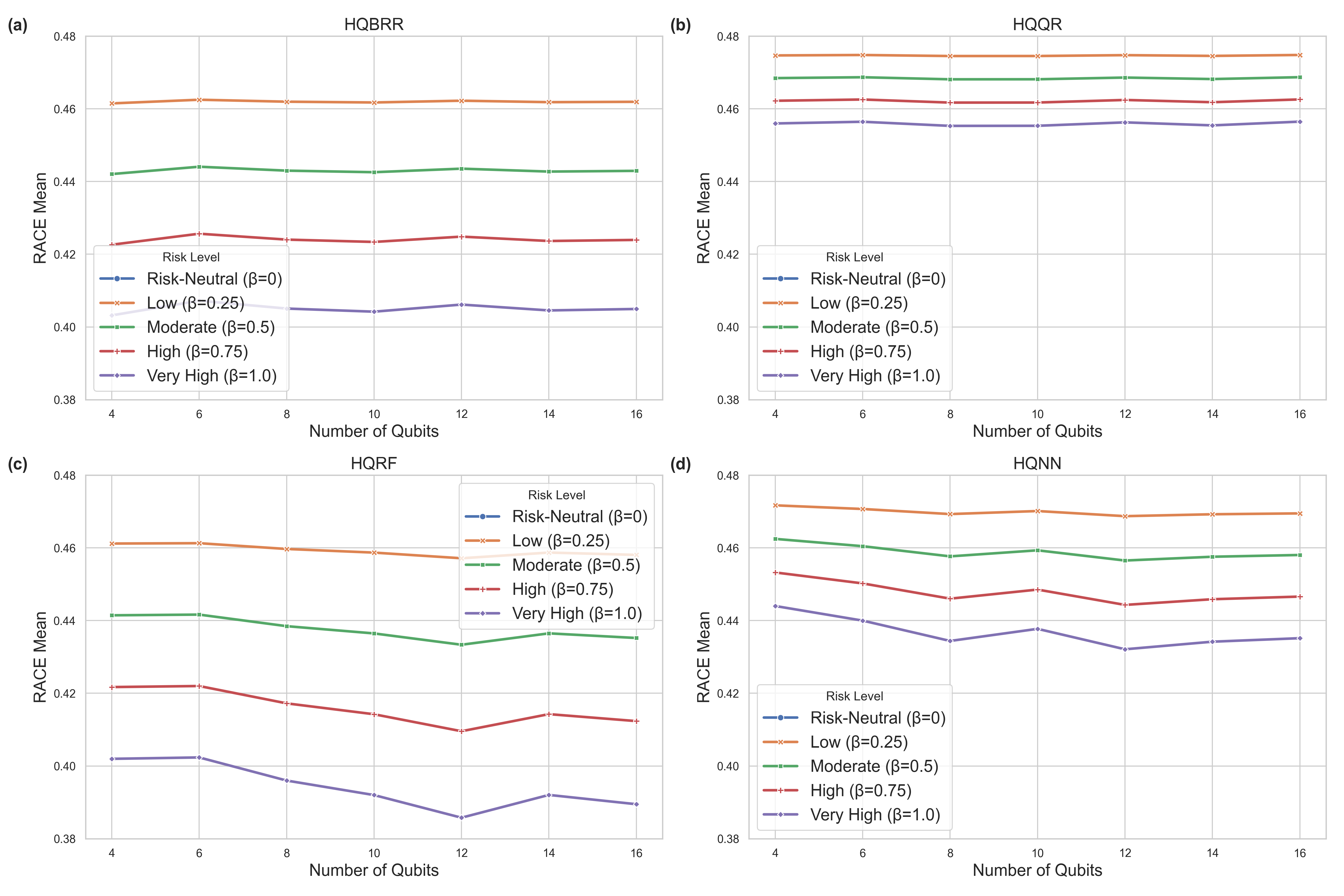}
  \caption{RACE variation with increasing qubit count for
quantum-classical hybrid models.}
    \label{fig:All_Models_Line_Plot_RACE_vs_Qubits_Subplots_Range}
    \footnotesize
    \vspace{\baselineskip}
\end{figure}

The effect of increasing qubits on RACE was evaluated for each model across all risk levels, as shown in Figure \ref{fig:All_Models_Line_Plot_RACE_vs_Qubits_Subplots_Range}. As qubits increase, the mean RACE exhibits notable variations in HQRF and HQNN, particularly at higher risk aversion levels ($\beta=0.75$ and $\beta=1.0$). For HQRF, a distinct dip in RACE occurs between 10 and 12 qubits, where values for $\beta=1.0$ drop to approximately 0.39 before a slight recovery at the 16th qubits. This suggests that the model becomes more sensitive to increasing qubits, potentially due to higher variability in decision trees with complex quantum states. Similarly, HQNN shows a decline in RACE between 10 and 12 qubits, particularly for higher $\beta$, with RACE dropping to around 0.43-0.44. This may indicate that NNs face challenges to perform consistently as qubit interactions grow more complex. In contrast, HQBRR and HQQR maintain near-constant RACE across qubit increases, highlighting their robustness to qubit scaling. The greater variability in HQRF and HQNN could be caused by the increased complexity of these models.

The experimental findings in this work confirm that quantum-classical hybrid models provide a difference in managing predictive uncertainty, particularly in complex, data-intensive environments. The HQBRR model exhibited consistently narrower uncertainty bounds compared to classical models, particularly as qubit counts increased. This change aligns with \cite{cerezo2021variational}, who noted the benefits of VQAs in predictive capabilities within hybrid models. Increasing qubit configurations from 4 to 16 showed that higher qubit counts enhance predictive uncertainty propagation, which strengthens the model's robustness and the detection of outlier data. To the best of the authors' knowledge, a similar UQ quantification analysis by increasing the qubits for comparison could not be found. However, the observation of this paper aligns with insights from work \cite{lloyd2020quantum}, \cite{adebiyi2023quantum}, and \cite{zaman2024comparative}, which demonstrated that larger quantum feature spaces enable better data representation and generalisation. It further suggests that adding quantum feature transformations to classical UQ frameworks offers a scalable way to study robustness in practical applications, such as supply chain resilience and financial risk assessment, where data variability is challenging. Models such as KACQ-DCNN \cite{ jahin2024kacqdcnnuncertaintyawareinterpretablekolmogorovarnold} and DiffHybrid-UQ \cite{akhare2023diffhybrid} further demonstrate the effectiveness of hybrid approaches in improving interpretability and reliability in UQ. The results in this work emphasise the value of quantum feature transformations and qubit scaling as practical tools for advancing UQ in quantum-classical hybrid models, showing their suitability for high-stakes data-driven settings.

\section{Conclusions}
\label{sec:conclusions}
This study investigated Uncertainty Quantification (UQ) within a quantum-classical hybrid machine learning framework focusing on high-stakes applications such as supply chain digital twins and financial risk assessment. By applying established UQ techniques to both classical and quantum-transformed features, it is demonstrated that quantum-classical hybrid models offer unique predictions compared to classical models alone in complex data environments. This experiment showed that increasing the qubit count from 4 to 16 shows significantly different uncertainties, which illustrates the effects of qubit scaling on UQ propagation. The key contributions of this study are (1) a systematic examination of how quantum feature transformations, facilitated through qubit scaling, impact UQ outcomes in hybrid models, (2) the introduction of qubit scaling as a method to improve outlier detection, showing that higher qubit configurations enhance model sensitivity to data variability, and (3) the findings can be extended to real-world problems of supply chain management, showing how combining quantum and classical computing methods can help make better decisions when dealing with complex, fast-changing situations where the stakes are high. This research provides insight for new studies on using quantum computing to better measure uncertainty, which could help companies make more accurate data-driven decisions, especially when working with large and complicated datasets.

In future, we aim to investigate the impact of qubit count on the performance of hybrid-quantum models using quantum circuits for direct prediction tasks rather than only for feature transformation. We also aim to identify any thresholds beyond which additional qubits no longer contribute to improving results. We plan to extend the testing of these UQ frameworks to more complex datasets. 


\bibliographystyle{IEEEtran}  
\bibliography{references}

\clearpage
\onecolumn

\section*{Supplementary files and data}

\section*{Synthetic data generation}
The synthetic data can be generated by following the below steps. 

\begin{enumerate}

\item Libraries: 
Import libraries such as \texttt{numpy}, \texttt{pandas}, and \texttt{sklearn} (for train/test split, pre-processing, classification, regression, etc.).

\item Seed: 
Set a random seed using \texttt{np.random.seed()} to ensure reproducibility of results.

\item Generate synthetic dataset:
   \begin{itemize}
      \item Create \texttt{product\_demand} using a Poisson distribution.
      \item Create features such as \texttt{inventory\_levels}, \texttt{lead\_time}, \texttt{order\_quantity}, \texttt{transportation\_costs}, \texttt{storage\_costs}, \texttt{production\_costs}, and \texttt{sales\_price} using normal distributions with different means and standard deviations.
      \item Generate features such as \texttt{supplier\_reliability}, \texttt{customer\_satisfaction}, \texttt{warehouse\_capacity\_utilisation}, etc., using uniform distributions.
      \item Use categorical variables for \texttt{weather\_conditions} and \texttt{market\_trends}.
      \item Generate other columns such as \texttt{supply\_chain\_disruption}, \texttt{economic\_shocks}, \texttt{natural\_disasters}, and \texttt{cybersecurity\_threats} using random choices and probabilities.
   \end{itemize}

\item Label encode categorical features: 
Convert the categorical features such as \texttt{weather\_conditions} and \texttt{market\_trends} into a numerical format using \texttt{LabelEncoder}.

\item Create a new feature for cost efficiency: 
Compute \texttt{cost\_efficiency} as a weighted sum of several cost features such as production, transportation, storage, and labour costs with added random noise:

\item Define the feature columns and target variable: 
\begin{itemize}
   \item Feature columns: All columns except \texttt{cost\_efficiency} and \texttt{disruption\_impact}.
   \item Target variable: \texttt{disruption\_impact}.
\end{itemize}

\item Split the dataset: 
Split the dataset into training and testing sets, allocating 70\% to training and 30\% to testing.

\item Standardise the feature columns: 
Use \texttt{StandardScaler} to normalise the feature columns:
   \begin{itemize}
      \item Fit the scaler on the training data.
      \item Transform both the training and testing data using the fitted scaler.
   \end{itemize}
\end{enumerate}

\begin{figure*}[!hb]
    \centering
    \includegraphics[width=\textwidth]{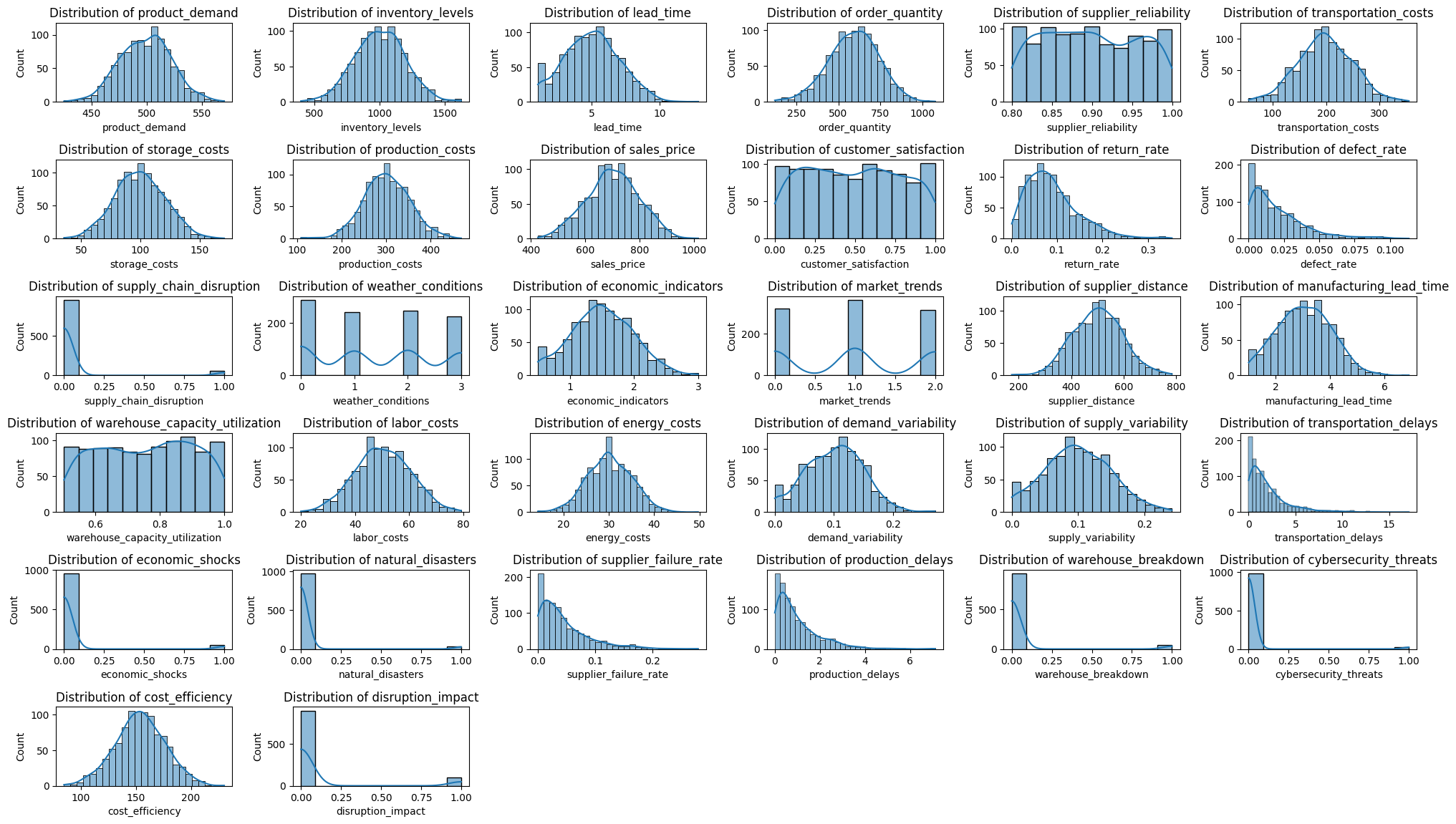}
    \caption{Data distribution for all features (synthetic data).}
    \label{fig:data_distribution}
    \footnotesize
\end{figure*}

A distribution of such generated data is given in Figure \ref{fig:data_distribution}.

\section*{Qubits Configuration}
\begin{figure*}[!ht]
    \centering
    \includegraphics[width=\textwidth]{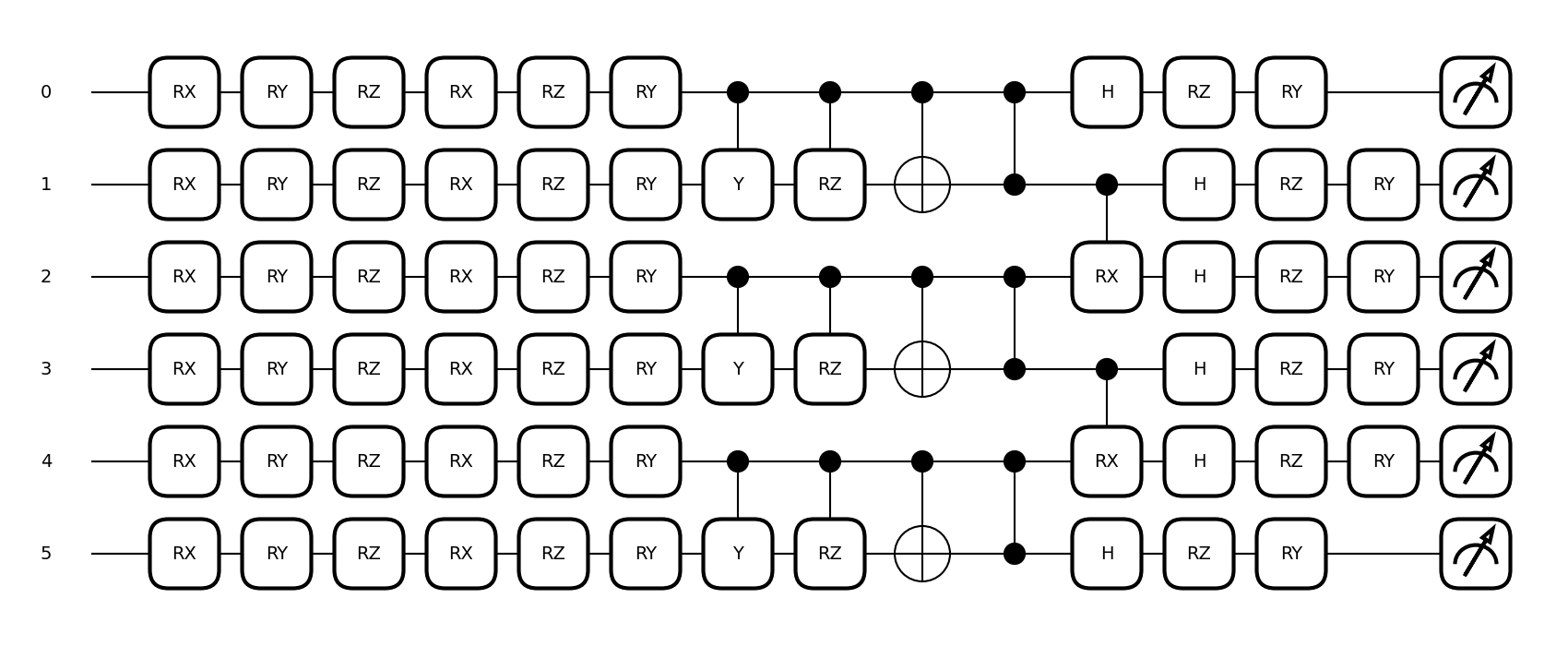}
    \captionsetup{justification=centering}  
    \caption{Circuit diagram for 6 qubits configuration.}
    \label{fig:quantum_circuit_6_qubits}
    \footnotesize
\end{figure*}

\begin{figure*}[!ht]
    \centering
    \includegraphics[width=\textwidth]{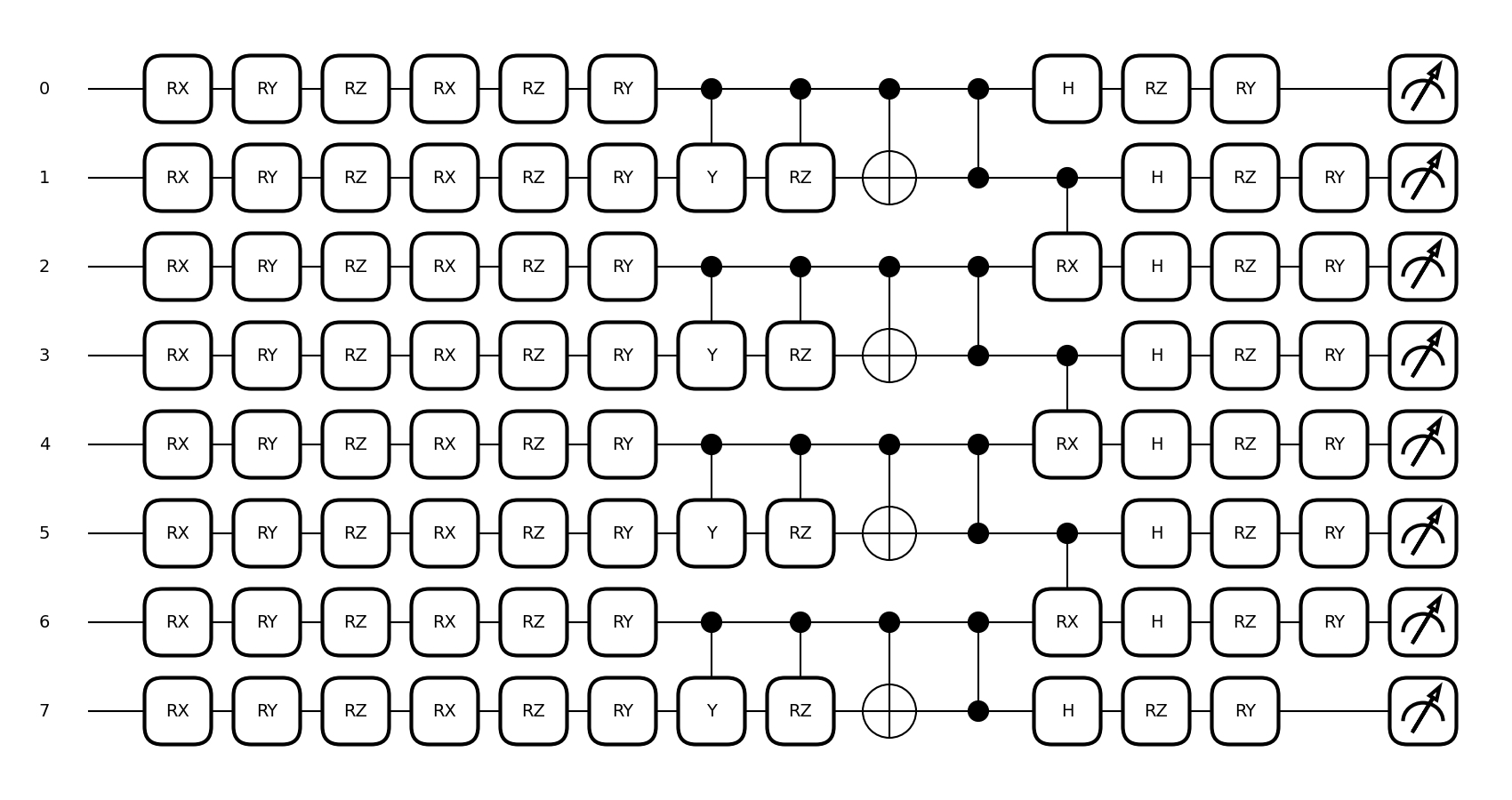}
    \caption{Circuit diagram for 8 qubits configuration.}
    \label{fig:quantum_circuit_8_qubits}
    \footnotesize

\end{figure*}

\begin{figure*}[!ht]
    \centering
    \includegraphics[width=\textwidth]{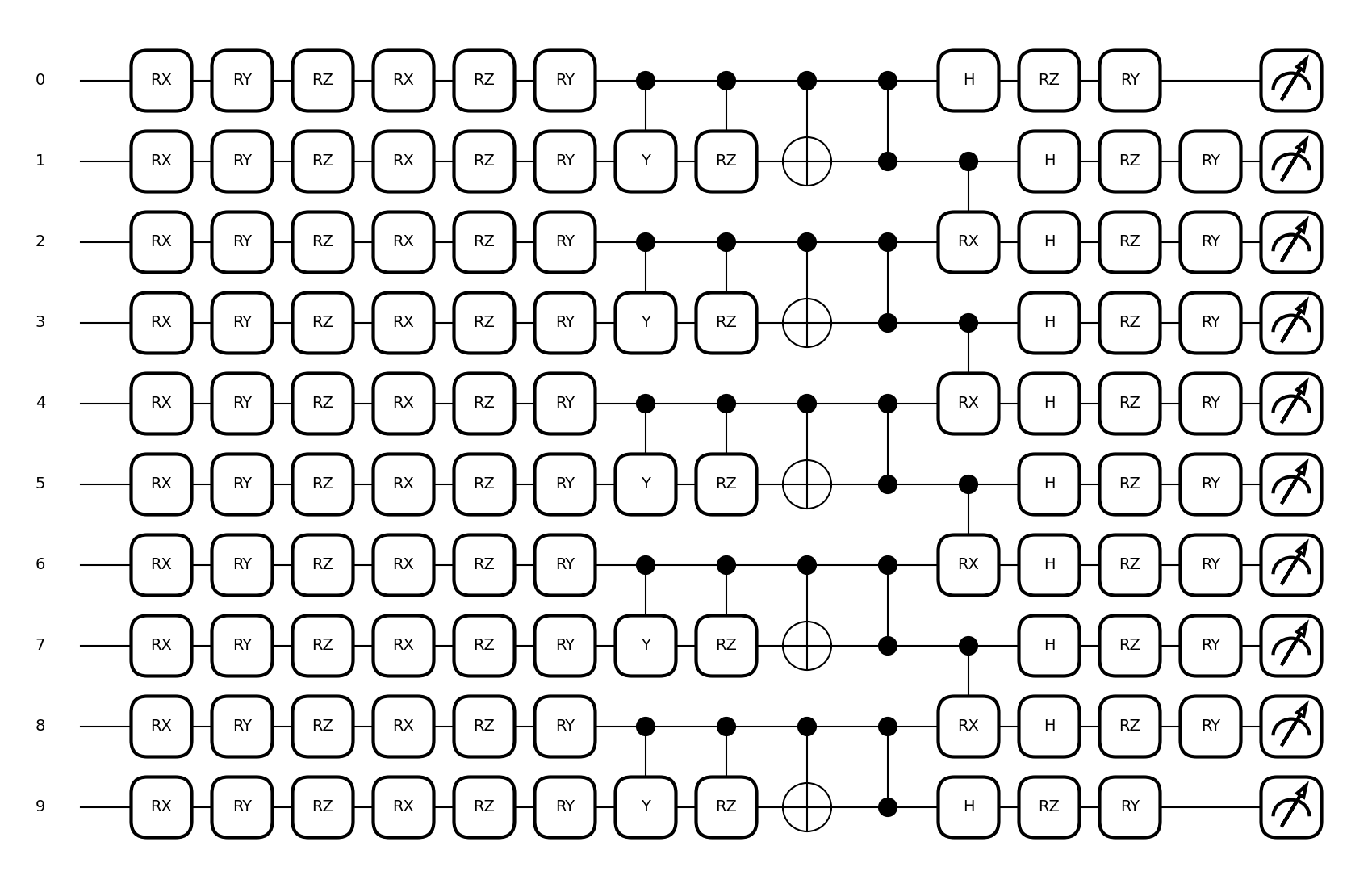}
    \caption{Circuit diagram for 10 qubits configuration.}
    \label{fig:quantum_circuit_10_qubits}
    \footnotesize

\end{figure*}

\begin{figure*}[!ht]
    \centering
    \includegraphics[width=\textwidth]{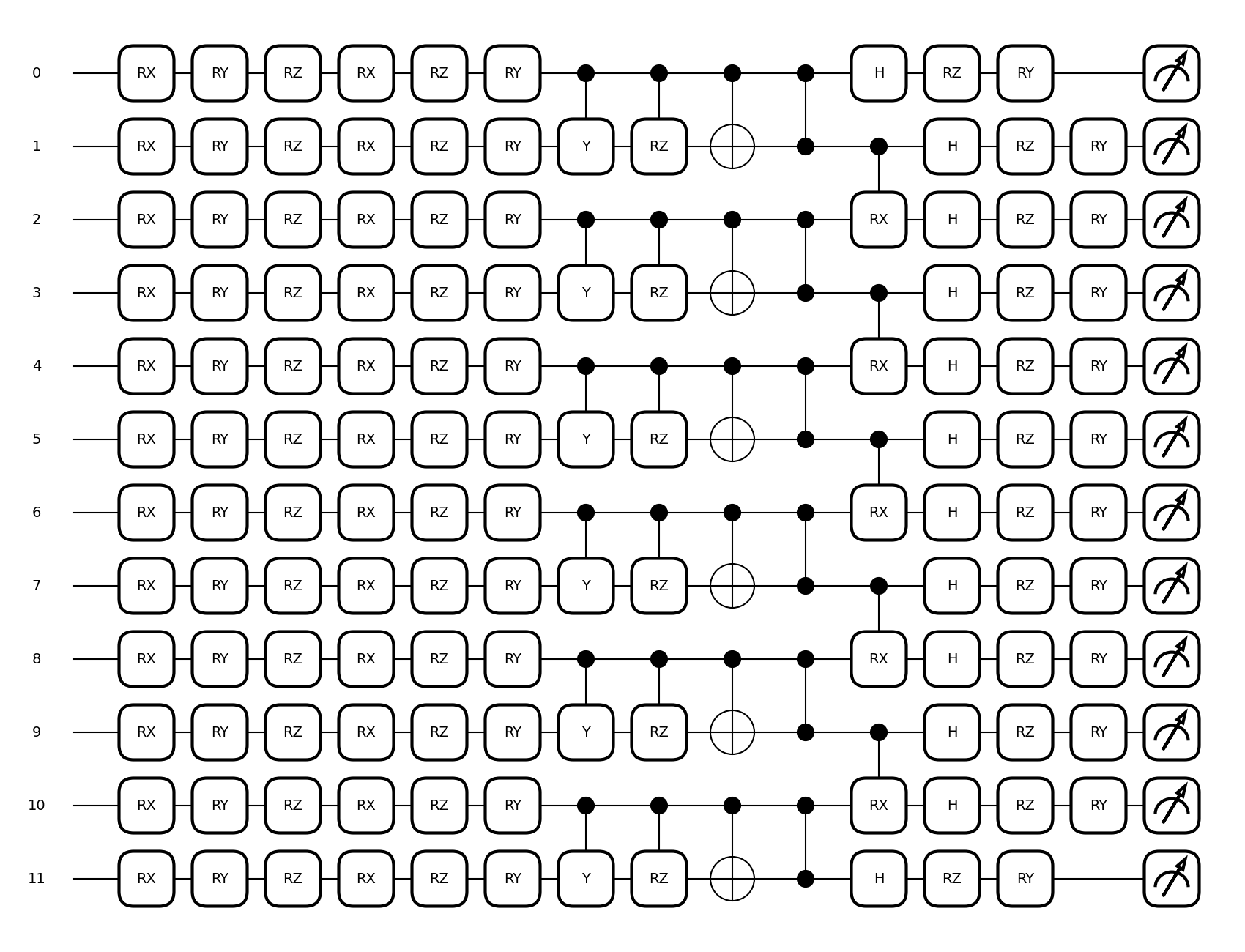}
    \caption{Circuit diagram for 12 qubits configuration.}
    \label{fig:quantum_circuit_12_qubits}
    \footnotesize

\end{figure*}

\begin{figure*}[!ht]
    \centering
    \includegraphics[width=\textwidth]{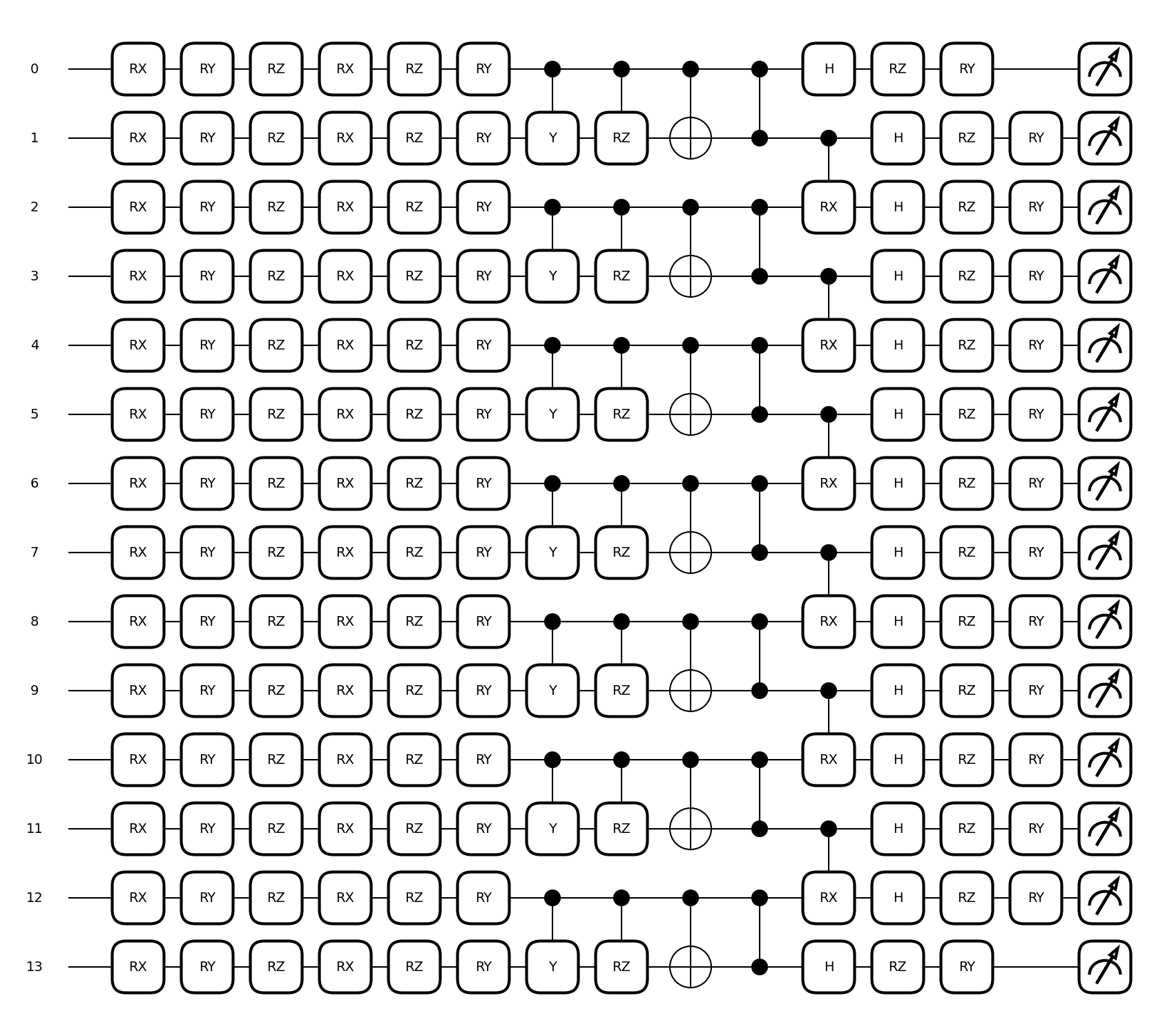}
    \caption{Circuit diagram for 14 qubits configuration.}
    \label{fig:quantum_circuit_14_qubits}
    \footnotesize

\end{figure*}

\begin{figure*}[!ht]
    \centering
    \includegraphics[width=\textwidth]{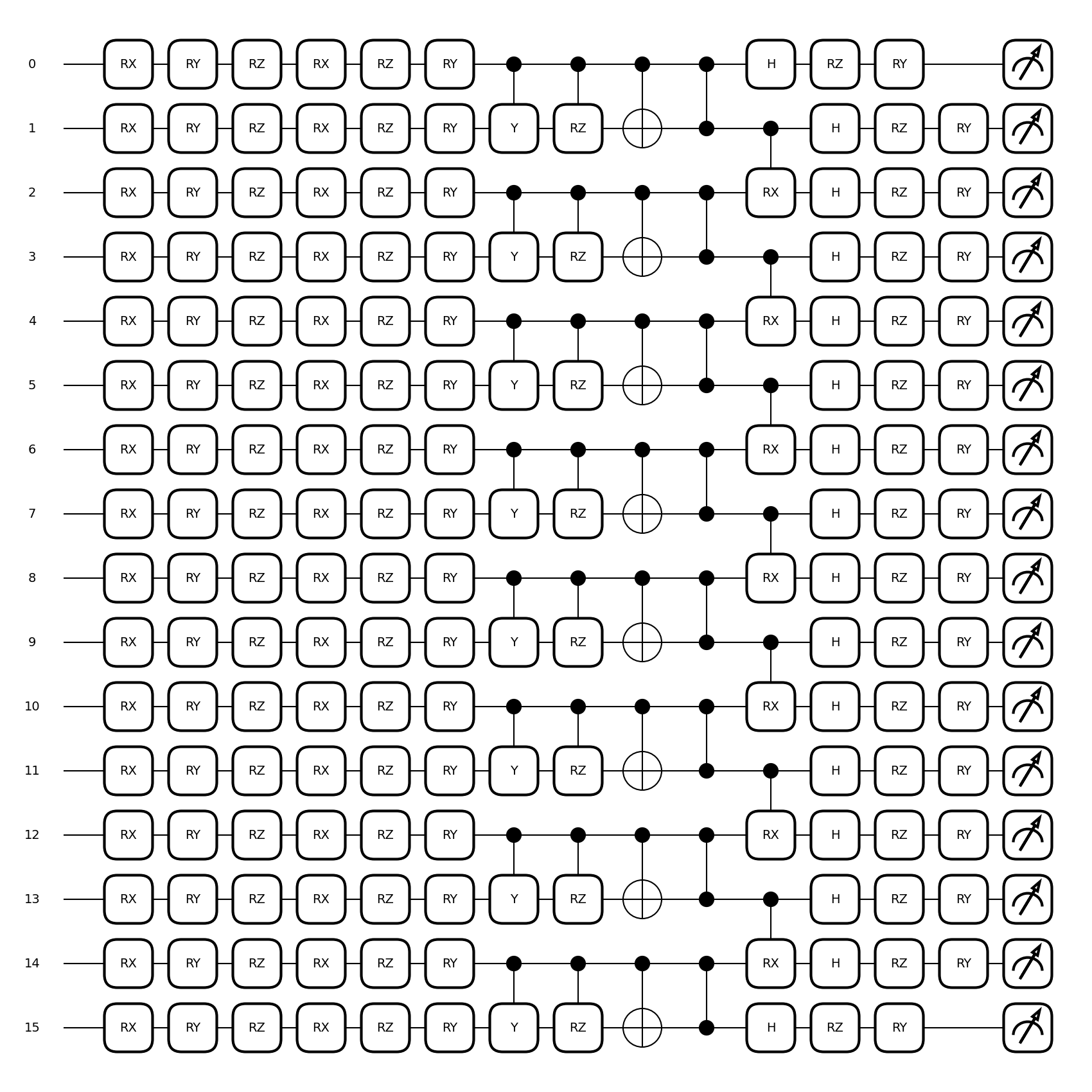}
    \caption{Circuit diagram for 16 qubits configuration.}
    \label{fig:quantum_circuit_16_qubits}
    \footnotesize

\end{figure*}

The default circuit chosen for the experiment was 4 qubits. The rest of the circuits are given in Figures \ref{fig:quantum_circuit_6_qubits}, \ref{fig:quantum_circuit_8_qubits}, \ref{fig:quantum_circuit_10_qubits}, \ref{fig:quantum_circuit_12_qubits}, \ref{fig:quantum_circuit_14_qubits} and \ref{fig:quantum_circuit_16_qubits}. 


\end{document}